\documentclass[10pt,twocolumn,letterpaper]{article}
\pdfoutput=1

\usepackage{cvpr}
\usepackage{times}
\usepackage{epsfig}
\usepackage{graphicx}
\usepackage{amsmath}
\usepackage{amssymb}

\usepackage{listings}
\usepackage{float}
\usepackage{color}
\usepackage{xcolor}

\lstdefinestyle{customc}{
  belowcaptionskip=1\baselineskip,
  breaklines=true,
  frame=L,
  xleftmargin=\parindent,
  language=C,
  showstringspaces=false,
  basicstyle=\footnotesize\ttfamily,
  keywordstyle=\bfseries\color{green!40!black},
  commentstyle=\itshape\color{purple!40!black},
  identifierstyle=\color{blue},
  stringstyle=\color{orange},
}

\usepackage[breaklinks=true,bookmarks=false]{hyperref}

\cvprfinalcopy 

\def\cvprPaperID{ECV \#18} 

\begin{document}

\title{The Indirect Convolution Algorithm}

\author{Marat Dukhan\\
Google Research\\
1600 Amphitheatre Parkway,\\
Mountain View, CA, 94043\\
{\tt\small maratek@google.com}
}

\maketitle

\begin{abstract}
   Deep learning frameworks commonly implement convolution operators with GEMM-based algorithms. In these algorithms, convolution is implemented on top of matrix-matrix multiplication (GEMM) functions, provided by highly optimized BLAS libraries. Convolutions with 1x1 kernels can be directly represented as a GEMM call, but convolutions with larger kernels require a special memory layout transformation - im2col or im2row - to fit into GEMM interface.
   
   The Indirect Convolution algorithm provides the efficiency of the GEMM primitive without the overhead of im2col transformation. In contrast to GEMM-based algorithms, the Indirect Convolution does not reshuffle the data to fit into the GEMM primitive but introduces an indirection buffer --- a buffer of pointers to the start of each row of image pixels. This broadens the application of our modified GEMM function to convolutions with arbitrary kernel size, padding, stride, and dilation.
   
   The Indirect Convolution algorithm reduces memory overhead proportionally to the number of input channels and outperforms the GEMM-based algorithm by up to 62\% on convolution parameters which involve im2col transformations in GEMM-based algorithms. This, however, comes at cost of minor performance reduction on 1x1 stride-1 convolutions.
\end{abstract}

\section{Introduction}

Convolutional neural networks (CNNs) are state-of-the-art models for many computer vision problems, including image classification \cite{ResNeXt, SENet, NASNet, FBNet}, object detection \cite{YoloV3, RetinaNet, PoolingPyramidNetwork}, semantic \cite{DeepLabV3Plus}, instance \cite{MaskRCNN}, and panoptic \cite{PanopticFPN} segmentation, image super-resolution, and denoising. Outside of computer vision, CNNs are used in speech synthesis \cite{WaveNet} and Natural Language Processing \cite{GatedCNN, DynamicConvolutions}. Typically, between 50-90\% of inference time in CNN models is spent in Convolution and closely related Transposed Convolution operators \cite{ShuffleNetV2}.

Optimal implementations of convolution operators are an active area of research. We briefly overview existing algorithms below for the case of 2D Convolution operator:

\begin{lstlisting}[language=C, style=customc, caption=Implementation of Direct Convolution Algorithm as 7 nested loops, label=lst:direct-convolution]
for (int n = 0; n < N; n++)
 for (int oy = 0; oy < Hout; oy++)
  for (int ox = 0; ox < Wout; ox++)
   for (int oc = 0; oc < K; oc++)
    for (int ky = 0; ky < R; ky++)
     for (int kx = 0; kx < S; kx++)
      for (int ic = 0; ic < C; ic++) {
       const int iy = oy * SY + ky * DY - PL;
       const int ix = ox * SX + kx * DX - PT;
       if (0 <= ix < Win && 0 <= iy < Hin)
        output[n][oy][ox][oc] +=
         input[n][iy][ix][ic] *
         filter[oc][ky][kx][ic];
      }
\end{lstlisting}

\textbf{The direct convolution algorithm} in its simplest form is expressed as 7 nested loops iterating along batch size $N$, output height $\mathit{H_{out}}$, output width $\mathit{W_{out}}$, output channels $\mathit{K}$, and accumulating partial results across kernel height $\mathit{R}$, kernel width $\mathit{S}$, and input channels $\mathit{C}$, as illustrated in Listing \ref{lst:direct-convolution}. Practical implementations of the direct convolution algorithm usually add extra loops for cache blocking and vectorization, which further complicate computation flow. Due to the large number of parameters and loops it is infeasible to write a single implementation that would deliver good performance across all parameter values with a standard layout for input and output tensors. Thus, deep learning frameworks and libraries which implement direct convolution algorithm adapted several strategies to cope with code multi-versioning complexity.

The most common strategy, implemented \eg in cuDNN \cite{cuDNN} and HexagonNN libraries, as well as MACE \cite{MACE} and ncnn frameworks, is to use an optimized direct convolution implementation for the most popular convolution parameters, such as 5x1, 1x5 and 3x3 stride-2 kernels, and fall back to a default algorithm for non-common values.

As another strategy, Intel MKL-DNN \cite{AnatomyOfHighPerformanceConvolution} and Intel LibXSMM \cite{LibXSMM} libraries employ parametrized architecture-specific Just-in-Time code generator to produce optimized direct convolution implementation for parameters provided at runtime. TVM \cite{TVM} takes the code-generation approach even further by combining Just-in-Time code generation with auto-tuning.

Several works \cite{HighPerformanceDirectConvolution, AnatomyOfHighPerformanceConvolution} suggest that direct convolution performance can be greatly improved through the use of a specialized processor-specific layouts. However, practical applications of this approach requires either a large set of neural network operations implemented for such specialized layout, or costly layout conversions for convolution inputs and outputs.

\textbf{GEMM-based algorithms} express computations in the Convolution operator as a GEMM (Matrix-Matrix Multiplication) operation. Such reformulation of the problem enables convolution implementations to leverage highly optimized BLAS (Linear Algebra) libraries, which exists for nearly every platform, and benefit from decades of research on efficient dense linear algebra computations \cite{GotoBLAS, BLIS}. GEMM-based algorithms can support arbitrary parameters, and are well-suited for a generic implementation of Convolution operator. As a result, the GEMM-based algorithm, introduced by Chellapilla et al \cite{ConvNetForDocumentProcessing}, are now used in all major deep learning frameworks, including TensorFlow \cite{TensorFlow}, PyTorch \cite{PyTorch}, and Caffe \cite{Caffe}.

GEMM-based algorithms rely on im2col or im2row memory transformations to convert the Convolution problem into a GEMM problem. These transformations copy a patch of input pixels that affect the value of an output pixel into a matrix row that corresponds to the output pixel. The product of this matrix of input patches and the corresponding filter tensor flattened into a matrix is an output matrix where rows represent output pixels and columns represent output channels. Building the patch matrix involves non-trivial overhead in memory storage and bandwidth. This overhead is proportional to kernel size, \eg for 3x3 convolution every pixel of the input matrix is replicated 9 times. Several modifications to the GEMM-based algorithms were proposed \cite{MemoryEfficientConvolution, LowMemoryGEMMBasedConvolution} that reduce the storage overhead by computing convolution in multiple small GEMM operations, each using a smaller patch matrix.

\textbf{Fast Convolution Algorithms} use Fourier \cite{FBFFT} or Winograd \cite{WinogradConvolution} transformations to reduce the computational complexity of Convolution with large kernel sizes. In Fast Convolution Algorithms the asymptotic complexity of Convolution operation does not depend on kernel size, and enables multiple-times speedups for Convolutions with large kernel sizes. Unfortunately, the algorithmic speedup is limited to specific Convolution parameters (large kernel sizes, unit stride and dilation, sufficiently large input size, and number of input and output channels), which prevents using Fast Convolution Algorithms as the default option.

\subsection{Contributions}

In this paper we present a novel type of algorithm for Convolution computation, named the \textbf{Indirect Convolution algorithm}. The Indirect Convolution algorithm is a modification of GEMM-based algorithms, and like GEMM-based algorithms it can efficiently support arbitrary Convolution parameters, and leverage the vast trove of research on high-performance GEMM implementation. Additionally, the Indirect Convolution algorithm has two major advantages over GEMM-based algorithms:
\begin{itemize}
\setlength{\parskip}{0pt}
\setlength{\itemsep}{0pt plus 1pt}
\item The Indirect Convolution algorithm \textbf{eliminates expensive and memory-intensive im2col transformations}. Elimination of im2col transformations improves performance by up to 62\% compared to GEMM-based algorithms. The performance improvement is particularly prominent on Convolutions with small number of output channels, when im2col comprises a large share of Convolution runtime.

\item The Indirect Convolution algorithm allows to \textbf{replace the im2col buffer with a much smaller indirection buffer}. The size of im2col buffer scales linearly with the number of input channels, but the size of indirection buffer does not depend on the number of input channels. Thus, the memory footprint advantage of Indirect Convolution algorithm is the greatest for Convolutions with many input channels.

\end{itemize}

\subsection{Limitations}

The high efficiency of Indirect Convolution algorithm is contingent on certain conditions:

\begin{itemize}
\setlength{\parskip}{0pt}
\setlength{\itemsep}{0pt plus 1pt}
\item The algorithm is optimized for \textbf{NHWC layout}, supported by TensorFlow \cite{TensorFlow}, TensorFlow Lite, and Caffe2 frameworks. While the algorithm can be adapted to work in NCHW layout (native to PyTorch \cite{PyTorch} and Caffe \cite{Caffe} frameworks), we don't expect it would be competitive with state-of-the-art patch-building algorithms of Andersen et al \cite{HighPerformanceDirectConvolution} due to strided memory access.

\item The algorithm is optimized for the forward pass of Convolution operator, and has \textbf{limited applicability to backward pass of Convolution operator and to the Transposed Convolution operator}. The Indirect Convolution algorithm presented in this paper replaces the GEMM-based algorithms using im2col or im2row transformations. However, for the backward pass of a strided Convolution operator and for strided Transposed Convolution operator, col2im and row2im-based algorithms are more optimal due to smaller number of arithmetic operations.

\item Similarly to GEMM-based convolution, the \textbf{Indirect Convolution algorithm is not efficient for depthwise convolutions}. Depthwise convolutions \cite{Xception} independently convolve each channel with its own set of filters.

\end{itemize}

\section{The Indirect Convolution algorithm}

The Indirect Convolution algorithm can be represented as a modification of GEMM-based algorithms. Where GEMM-based algorithms reshuffle data to fit it into GEMM interface, the Indirect Convolution Algorithm instead modifies the GEMM primitive to adopt it to the original data layout. The modified GEMM primitive, denoted the \textbf{Indirect GEMM} in this paper, has a similar computational structure as the standard GEMM, and can reuse the same optimizations.

\subsection{GEMM Primitive}

\begin{figure}[t]
\centering
 \includegraphics[width=0.8\linewidth]{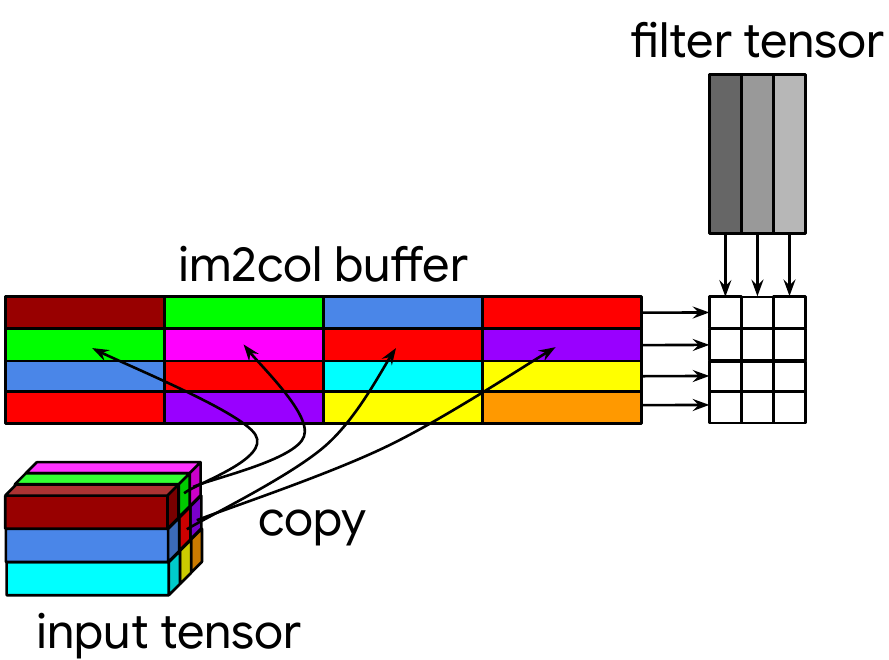}
\caption{GEMM operation as a component of GEMM-based convolution algorithm. im2col buffer represents matrix A, filter tensor - matrix B, and their product constitutes the output tensor.}
\label{fig:gemm}
\end{figure}

For an $M \times K$ matrix $A$ (optionally transposed), $K \times N$ matrix $B$ (optionally transposed), and $M \times N$ matrix C, and scalar constants $\alpha$ and $\beta$, the GEMM primitive computes

\[C \leftarrow \alpha A \times B + \beta C \]

In the context of the forward pass of a Convolution operator, $A$ contains input tensor data, $B$ the constant filter data, and $C$ represents output tensor data. In the traditional im2col+GEMM algorithm $\alpha = 1$, and $\beta = 0$, albeit newer low-memory GEMM-based algorithms \cite{LowMemoryGEMMBasedConvolution} make use of $\beta = 1$ case as well. Fig. \ref{fig:gemm} illustrates the role of GEMM primitive in GEMM-based convolution algorithm, and Listing \ref{lst:gemm-microkernel} demonstrates the basic building block of a GEMM primitive -- a GEMM micro-kernel that produce 2 rows and 2 columns of matrix C.

\begin{lstlisting}[language=C, style=customc, caption=Implementation of GEMM micro-kernel in C, label=lst:gemm-microkernel]
void uGEMM(
 int k, const float* pw,
 const float* pa, int lda,
 float* pc, int ldc)
{
 const float* pa0 = pa;
 const float* pa1 = pa + lda;
 float c00 = 0, c01 = 0, c10 = 0, c11 = 0;
 do {
  const float a0 = *pa0++, a1 = *pa1++;
  const float b0 = *pw++;
  c00 += a0 * b0;
  c10 += a1 * b0;
  const float b1 = *pw++;
  c01 += a0 * b1;
  c11 += a1 * b1;
 } while (--k != 0);
 pc[0] = c00; pc[1] = c01;
 pc += ldc;
 pc[0] = c10; pc[1] = c11;
}
\end{lstlisting}

\subsection{From GEMM to Indirect GEMM}

\begin{figure}[t]
\centering
 \includegraphics[width=0.8\linewidth]{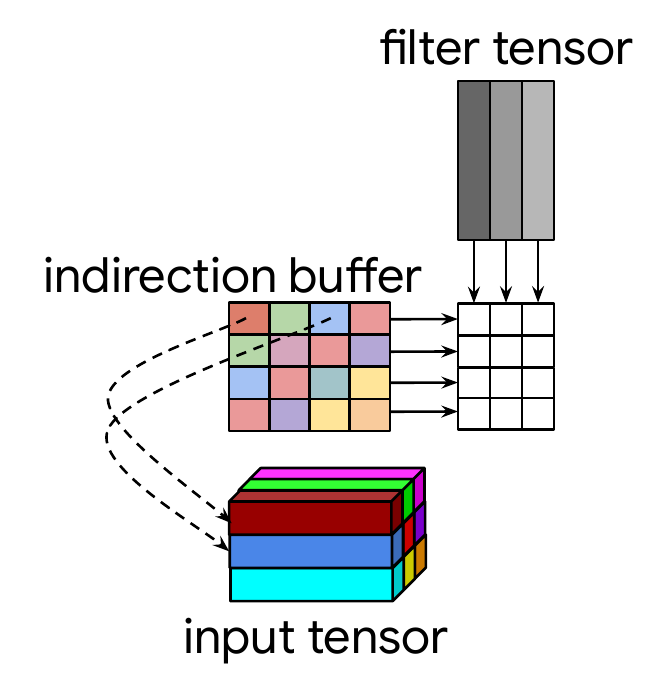}
\caption{Indirect GEMM operation as a component of Indirect Convolution algorithm. The indirection buffer contains only pointers to rows of the input tensor, and the Indirect GEMM operation reads rows of data directly from the input tensor.}
\label{fig:indirect-gemm}
\end{figure}

We suggest two modifications that jointly make the GEMM primitive directly suitable for the convolution implementation:

\begin{enumerate}
\setlength{\parskip}{0pt}
\setlength{\itemsep}{0pt plus 1pt}
\item Removing the assumption that rows of matrix A are separated in memory by a constant stride. Instead, pointers to rows of matrix A are loaded from an array of pointers provided by the caller and denoted \textbf{indirection buffer}. In the context of a 2D convolution, the indirection buffer specifies the address of a row of pixels in the input tensor that contribute to the computation of the output pixel.

\item Secondly, we add an extra loop over elements of the kernel. For each iteration of this loop, the modified GEMM primitive loads new pointers to input rows from the indirection buffer, computes dot products of $K$ elements specified by these pointers with the filter data, and accumulates the results of the dot product with results of the previous loop iterations.
\end{enumerate}

These modifications enable Indirect Convolution algorithm to implement a fused im2col + GEMM operation, but without ever materializing the result of im2col operation in memory. Fig. \ref{fig:indirect-gemm} illustrates the data flow in Indirect GEMM primitive, and Listing \ref{lst:indirect-gemm-microkernel} provides an example of an Indirect GEMM micro-kernel.

\begin{lstlisting}[language=C, style=customc, caption=Implementation of Indirect GEMM micro-kernel in C, label=lst:indirect-gemm-microkernel]
void uIndirectGEMM(
 int n, int k, const float* pw,
 const float** ppa, int lda,
 float* pc, int ldc)
{
 float c00 = 0, c01 = 0, c10 = 0, c11 = 0;
 do {
  const float *pa0 = *ppa++;
  const float *pa1 = *ppa++;
  int kk = k;
  do {
   const float a0 = *pa0++, a1 = *pa1++;
   const float b0 = *pw++;
   c00 += a0 * b0;
   c10 += a1 * b0;
   const float b1 = *pw++;
   c01 += a0 * b1;
   c11 += a1 * b1;
  } while (--kk != 0);
 } while (--n != 0);
 pc[0] = c00; pc[1] = c01;
 pc += ldc;
 pc[0] = c10; pc[1] = c11;
}
\end{lstlisting}

\subsection{Indirection Buffer}

The Indirection buffer is a buffer of pointers to rows of input pixels. Each row has $C$ pixels, and the rows can optionally be strided. For each output pixel position and for each kernel element the indirection buffer contains a pointer to a row of input pixels that would be convolved with a row of the filter weights for the corresponding kernel element to produce the corresponding output pixel.

It is common to use an implicit padding for convolutions with non-unit kernels. In convolutions with an implicit padding, the input tensor is implicitly padded with zeros along the spatial dimensions before computing convolution. To handle the padded convolution, the Indirect Convolution algorithm requires an explicit zero vector - a constant vector vector with $C$ elements initialized to zeros. The explicit zero vector does not need to be contigious with the input tensor, and can even be shared between multiple convolution operators. During an initializing of the indirection buffer, pointers to input rows which fall outside of the input tensor range are replaced with pointers to the explicit zero vector.

The Indirection buffer depends on several parameters: shapes of input, output, and filter tensors, convolution stride, dilation, and implicit padding, and pointers to input tensor and explicit zero tensor, and stride of pixel rows in the input tensors. These parameters can be categorized into several groups, according to the frequency of their change and implications of their change on indirection buffer:

\begin{itemize}
\setlength{\parskip}{0pt}
\setlength{\itemsep}{0pt plus 1pt}
    \item Convolution stride, dilation, kernel size, implicit padding, number of input channels, and output channels are parameters of a neural network model and once the model is instantiated, they are practically immutable.
    \item Changes in the height and width of input or output tensors require a complete reinitialization of indirection buffer. However, for most types of models, and in particular in the production environment, such changes are rare.
    \item Changes in the batch size require a partial reinitialization of the indirection buffer only for batch indices that were not previously initialized.
    \item Changes in pointers to the input tensor or the explicit zero vector require a complete reinitialization of indirection buffer. To avoid the cost, a high-level framework implementing the convolution can guarantee that in the absence of shape changes, tensors have persistent location.
\end{itemize}

\section{Experimental Evaluation}

Four factors affect performance of the Indirect Convolution compared to GEMM-based convolution algorithms:

\begin{enumerate}
\setlength{\parskip}{0pt}
\setlength{\itemsep}{0pt plus 1pt}
    \item Elimination of \textit{im2col} transformation for non-unit convolutions.
    \item Improved caching of the input rows for convolutions with large kernels as Indirect GEMM reads input rows contributing to different output pixels from the same location while GEMM would read these input rows from different locations in im2col buffer.
    \item Overhead of loading pointers to rows of the input data from indirection buffer compared to computing them under constant stride assumption.
    \item Potentially lower efficiency of two nested loops with $R \times S$ and $C$ iterations accordingly in the Indirect GEMM operation compared to a single loop with $R \times S \times C$ iterations in the GEMM operation.
\end{enumerate}

These factors are closely coupled together, but we can separately benchmark the effect of the first two (which positively affect the Indirect Convolution algorithm performance) and the last two (with negative effect on performance) by benchmarking three variants of convolution implementation:

\begin{itemize}
\setlength{\parskip}{0pt}
\setlength{\itemsep}{0pt plus 1pt}
    \item The Indirect Convolution algorithm
    \item The traditional GEMM-based Algorithm. Unless the convolution uses 1x1 kernel and unit stride, this algorithm involves im2col transformation.
    \item The GEMM part of the traditional GEMM-based algorithm. This benchmark excludes im2col transformation, and therefore does not produce the correct result. We include it only as a way to separate the effect of different factors on performance.
\end{itemize}

\subsection{Experimental Setup}

\begin{table}
\begin{center}
\begin{tabular}{|l|c|c|}
\hline
Device & Chipset & uArch \\
\hline\hline
Samsung Galaxy S8 & Exynos 8895    & Exynos-M2 \\
Google Pixel 2 XL & Snapdragon 835 & Cortex-A73 \\
Google Pixel 3    & Snapdragon 845 & Cortex-A75 \\
\hline
\end{tabular}
\end{center}
\caption{Characteristics of mobile devices in performance evaluation. Microarchitecture (uArch) is specified for the big cores.}
\label{table:devices}
\end{table}

\textbf{Platforms} We evaluated performance on three ARM64 Android devices with characteristics listed in Table \ref{table:devices}. The processors in these mobile devices include two types of cores: high-performance (big) cores and low-power (little) cores. In our experiments, all benchmarks were run in single-threaded mode with thread pinning to a single big core.

\textbf{Implementation} We use highly optimized implementations of GEMM and the Indirect GEMM micro-kernels in ARM64 assembly with software pipelining for out-of-order cores. Both micro-kernels produce 4x8 output tile (i.e. 4 output pixels with 8 output channels each), and use exactly the same instruction sequence in the inner loop of the Indirect GEMM micro-kernel and the main loop of GEMM micro-kernel.

Unlike many other GEMM implementations which use the Goto algorithm \cite{GotoBLAS}, we do not repack panels of matrix A accessed in a micro-kernel into a contiguous memory region. Goto and Van de Geijn \cite{GotoBLAS} suggested repacking as a way to overcome limited cache associativity. In contrast, we find that with GEMM matrices that typically occur in neural network architectures, limited cache associativity is not a concern because whole panels of A and B matrices read in the micro-kernel fit into level-1 cache. Note that avoiding repacking of matrix A in GEMM and Indirect GEMM primitives is our implementation detail, and both GEMM-based algorithm andthe  Indirect Convolution algorithm can be implemented either with or without repacking of inputs. However, our implementations of GEMM and Indirect GEMM micro-kernels assume that matrix B, which contains filter weights, is repacked into a contigious memory region, because filter weights never change at inference time, and such repacking can be done only once with no run-time cost.

\textbf{Protocol} We implement all micro-benchmarks on top of the Google Benchmark framework, which takes care of estimating sustained performance for the micro-benchmark. On top of it, each micro-benchmark is repeated 25 times. To bring measurements with different convolution parameters to a common scale, we compute resulting performance (in GFLOPS), and report median metric of the 25 runs, as well as 20\% and 80\% quantiles.

For each run of micro-benchmark we simulate the cache state during neural network inference: filter, bias, and output tensors, and indirection buffer are cleaned from cache, input tensor is prefetched into L1 cache, and im2col buffer stays in cache between convolution invocations to represent the same im2col buffer space re-used between different convolution operators. The indirection buffer is initialized only once (outside of the benchmarked snippet), and reused across invocations of the Indirect Convolution algorithm.

\begin{figure*}[!ht]
\centering
 \includegraphics[width=0.95\linewidth]{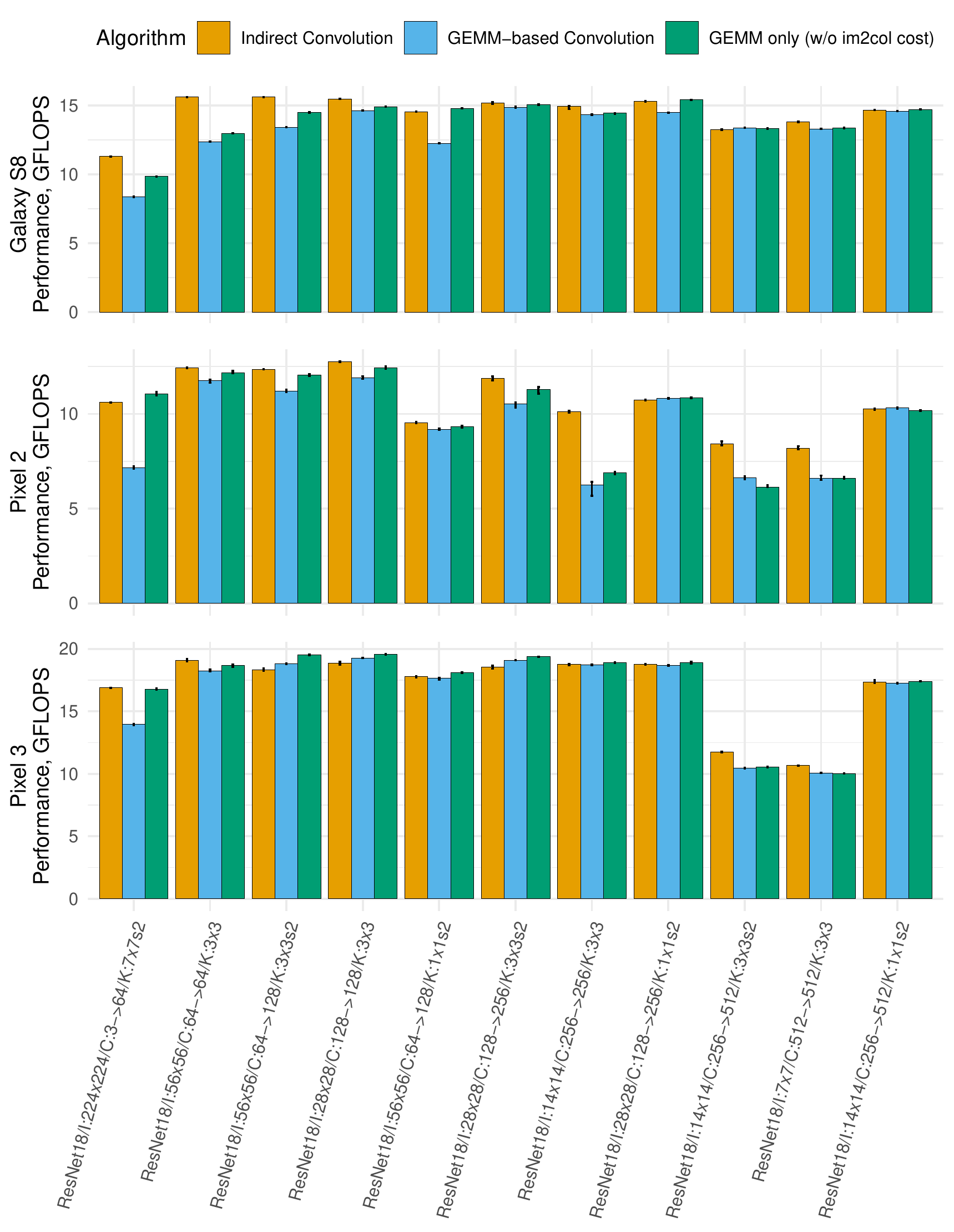}
\caption{Performance of the Indirect Convolution algorithm and GEMM-based Algorithm on convolution operators of the ResNet-18 model. Opaque bars represent median performance across 25 runs. Error bars represent 20\% and 80\% quantiles.}
\label{fig:resnet18}
\end{figure*}

\begin{figure*}[!ht]
\centering
 \includegraphics[width=0.95\linewidth]{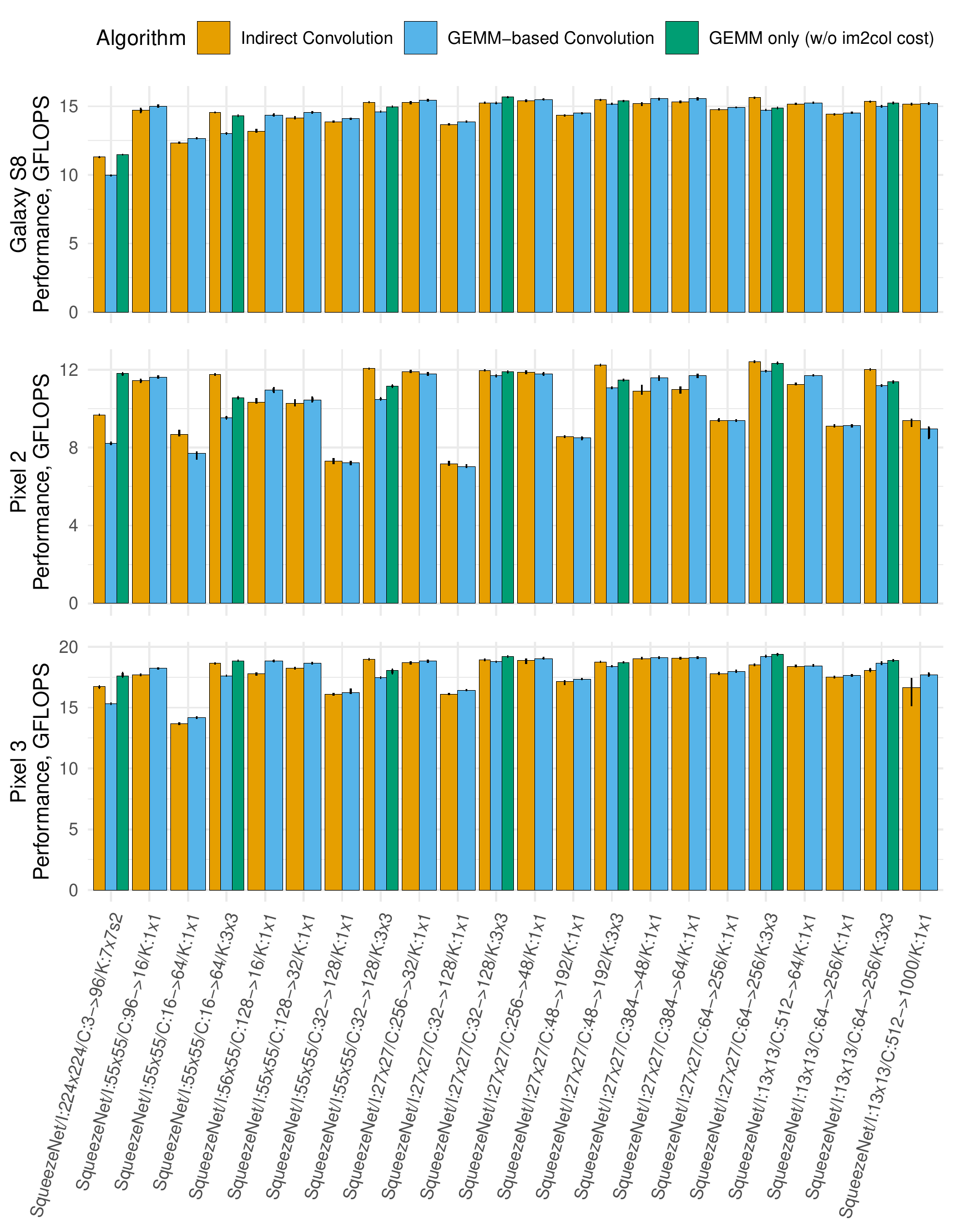}
\caption{Performance of the Indirect Convolution algorithm and GEMM-based Algorithm on convolution operators of the SqueezeNet 1.0 model. Opaque bars represent median performance across 25 runs. Error bars represent 20\% and 80\% quantiles.}
\label{fig:squeezenet}
\end{figure*}

\textbf{Models} We choose to evaluate performance on the convolution parameters of ResNet-18 \cite{ResNet} and SqueezeNet 1.0 \cite{SqueezeNet} models. Unlike more recent mobile-optimized models like MobileNet v2 \cite{MobileNetV2} and ShuffleNet v2 \cite{ShuffleNetV2}, which almost exclusively use 1x1 and depthwise convolutions, ResNet and SqueezeNet models employ the variety of convolution parameters, and provide a more balanced experimental workload. Both models start with a 7x7 stride-2 convolution, and then include 3x3 stride-1 convolutions. SqueezeNet additionally features 1x1 stride-1 convolutions and ResNet-18 makes use of 1x1 stride-2 and 3x3 stride-2 convolutions. Table \ref{table:models} summarized the composition of the convolution parameters in both models.

\begin{table}
\begin{center}
\begin{tabular}{|l|c|c|}
\hline
Convolution  & ResNet-18 & SqueezeNet 1.0 \\
\hline\hline
7x7 stride-2 & 1         & 1              \\
3x3 stride-2 & 3         & 0              \\
3x3 stride-1 & 4         & 6              \\
1x1 stride-2 & 3         & 0              \\
1x1 stride-1 & 0         & 15             \\
\hline
\end{tabular}
\end{center}
\caption{Types and count of Convolution operators in SqueezeNet 1.0 and ResNet-18 models. Convolutions with identical parameters are counted only once.}
\label{table:models}
\end{table}

\subsection{Experimental Results}

Fig. \ref{fig:resnet18} and \ref{fig:squeezenet} illustrate performance of the Indirect Convolution algorithm, GEMM-based algorithm, and just the GEMM part of the GEMM-based algorithm on the ResNet-18 and SqueezeNet 1.0 models respectively. In 1x1 stride-1 Convolutions in the SqueezeNet model the GEMM-based algorithm directly call into GEMM primitive without using im2col transformation; for this reason, we do not separately show GEMM-only performance for these Convolutions.

\begin{table}
\begin{center}
\begin{tabular}{|l|c|c|}
\hline
Device            &  non-1x1 & 1x1 stride-2 \\
\hline\hline
Samsung Galaxy S8 & +10.97\% & +8.02\% \\
Google Pixel 2 XL & +23.26\% & +0.84\% \\
Google Pixel 3    &  +4.31\% & +0.51\% \\
\hline
\end{tabular}
\end{center}
\caption{Geomean performance of modified GEMM primitive relative to standard GEMM primtive on 1x1 and non-1x1 Convolutions in ResNet-18 model.}
\label{table:resnet18}
\end{table}

\begin{table}
\begin{center}
\begin{tabular}{|l|c|c|}
\hline
Device            &  non-1x1 & 1x1 stride 1 \\
\hline\hline
Samsung Galaxy S8 &  +5.70\% & -1.84\%  \\
Google Pixel 2 XL & +11.29\% & -0.25\% \\
Google Pixel 3    &  +2.67\% & -1.91\%  \\
\hline
\end{tabular}
\end{center}
\caption{Geomean performance of modified GEMM primitive relative to standard GEMM primtive on 1x1 and non-1x1 Convolutions in SqueezeNet 1.0 model.}
\label{table:squeezenet}
\end{table}

These plots reveal that in most cases the Indirect GEMM has similar performance to the GEMM primitive post-im2col transformation; however, the addition of im2col transformation makes overall the Indirect Convolution algorithm outperform GEMM-based Convolution on all Convolutions which involve im2col transformation. Tables \ref{table:resnet18} and \ref{table:squeezenet} quantify this impact by types of convolution layers in ResNet-18 and SqueezeNet 1.0 models. The Indirect Convolution algorithm has varying impact depending on convolution parameters:

\begin{itemize}
\setlength{\parskip}{0pt}
\setlength{\itemsep}{0pt plus 1pt}
\item Convolutions with larger than 1x1 kernels see the biggest impact, with major performance improvement in $2.7-23.3\%$ range.
\item 1x1 stride-2 convolutions, which similarly need im2col transformation in GEMM-based algorithms, but don't benefit from improved cache locality in the Indirect GEMM, see smaller improvements in $0.5-8.0\%$ range.
\item 1x1 stride-1 convolutions, where GEMM-based algorithms incur no im2col overhead, demonstrate a minor performance regression in $0.3-1.9\%$ range, due to the extra complexity of the Indirect GEMM primitive compared to the tranditional GEMM.
\end{itemize}

These results suggest that the Indirect Convolution algorithm can provide substantial improvement for convolutions which involve im2col transformation in GEMM-based algorithms. However, the GEMM primitive has a small edge on 1x1 stride-1 convolutions, which do not involve im2col transformation, and for best overall performance it is beneficial to switch between GEMM or the Indirect GEMM primitives depending on convolution parameters.

\section{Analysis}

Roofline model \cite{Roofline} provides a convenient analysis tool for predicting which parameters affect the relative performance of the Indirect Convolution algorithm and GEMM-based algorithms. For the analysis, we denote output height and width as $H_{out}$ and $W_{out}$, input and output channels as $C$ and $K$, and kernel height and width as $R$ and $S$.

With the above notation, both GEMM and the Indirect GEMM algorithms involve $K \times C \times H_out \times W_out \times R \times S$ compute-bound operations (FLOPs)\footnote{We can exclude memory operations in GEMM and Indirect GEMM from our analysis, as these operations are almost always compute-bound.}. GEMM-based convolution with patch-building transformation additionally incurs $2 \times C \times H_{out} \times W_{out} \times R \times S$ memory operations. Assuming system's arithmetic intensity (ratio of FLOPs to memory loads in balanced code) $\lambda$, then we do $(K + 2\lambda) \times C \times H_{out} \times W_{out} \times  R \times S$ FLOPs-equivalent operations in GEMM-based convolution and the speedup of Indirect Convolution algorithm is $1 + \frac{2\lambda}{K}$. Thus, the Indirect Convolution algorithm is the most beneficial when the number of output channels is small. The upper bound on speedup is $1+2\lambda$, and suggests that as the systems' arithmetic intensity continues to grow with each generation, so will the advantage of Indirect Convolution algorithm.

\section{Conclusion}

The Indirect Convolution algorithm is a modification of GEMM-based Convolution algorithms where the GEMM operation reads addresses of rows in the input tensor from indirection buffer. Experiments revealed that this modified GEMM-like operation has similar performance as the traditional GEMM operation, and suggested that the major differences between the two types of algorithms stem from the difference between the im2col buffer in GEMM-based algorithms and the indirection buffer in the Indirect Convolution Algorithm. Unlike im2col buffer in GEMM-based algorithms, the Indirection buffer is constant in the number of input channels, and can persist between convolution invocations.

Indirect Convolution algorithm offers the universality of GEMM-based algorithm, but with smaller memory footprint and elimination of im2col transformation cost. These characteristics make the Indirect Convolution algorithm a viable option for default implementation of the convolution operator.

The Indirect Convolution algorithm potentially has interesting performance characteristics beyond the scope of this paper. In particular, the algorithm may have additional performance advantage over GEMM-based algorithms during multi-threaded convolution invocation. The modified GEMM operation is compute-bound, and should scale linearly with the number of cores, while im2col-component of GEMM-based convolution would be saturated by memory or cache bandwidth, which has sublinear scaling in number of cores. Exploring these other dimensions of performance is a topic for further research.

\section*{Acknowledgements}

We would like to thank Frank Barchard for contibuted optimizations to GEMM and Indirect GEMM micro-kernels, and Artsiom Ablavatski, Matthias Grundmann, Sergey Ioffe, and Juhyun Lee for helpful feedback on the paper. Indirect Convolution algorithm was previously implemented by the author in QNNPACK library \cite{QNNPACK}, and we are thankful to Bert Maher, Hao Lu, and Yiming Wu who contributed to this library.

{\small
\bibliographystyle{ieee_fullname}
\bibliography{indirect_convolution}
}

\end{document}